\documentclass[sn-mathphys,Numbered]{sn-jnl}

\usepackage{graphicx}%
\usepackage{multirow}%
\usepackage{amsmath,amssymb,amsfonts}%
\usepackage{amsthm}%
\usepackage{mathrsfs}%
\usepackage[title]{appendix}%
\usepackage{xcolor}%
\usepackage{textcomp}%
\usepackage{manyfoot}%
\usepackage{booktabs}%
\usepackage{algorithm}%
\usepackage{algorithmicx}%
\usepackage{algpseudocode}%
\usepackage{listings}%
\usepackage[justification=centering]{caption}%
\usepackage[T1]{fontenc}

\theoremstyle{thmstyleone}%
\theoremstyle{thmstyletwo}%

\theoremstyle{thmstylethree}%

\begin{document}

\title[Article Title]{HD-GCN:A Hybrid Diffusion Graph Convolutional Network}


\author[1]{\fnm{Zhi} \sur{Yang}}

\author[1]{\fnm{Kang} \sur{Li}}

\author*[1]{\fnm{Haitao} \sur{Gan}}\email{htgan01@hbut.edu.cn}

\author[1]{\fnm{Zhongwei} \sur{Huang}}

\author[1]{\fnm{Ming} \sur{Shi}}

\affil[1]{\orgdiv{School of Computer Science}, \orgname{Hubei University of Technology}, \city{Wuhan}, \postcode{430068}, \state{Hubei}, \country{China}}

\abstract{In tasks involving graph-structured data, Graph Convolutional Networks (GCNs) have brought promising results by introducing convolution into Graph Neural Networks (GNNs). However, the information diffusion performance of GCNs and its variant models is limited by the adjacency matrix, which can lower their performance. Therefore, we introduce a new framework for graph convolutional networks called the Hybrid Diffusion Graph Convolutional Network (HD-GCN) to address the limitations of information diffusion caused by the adjacency matrix. In the HD-GCN framework, we initially utilize diffusion maps to facilitate the diffusion of information among nodes that are adjacent to each other in the feature space. This allows for the diffusion of information between similar points that may not have an adjacent relationship. Next, we utilize graph convolution to further propagate information among adjacent nodes after the diffusion maps, thereby enabling the spread of information among similar nodes that are adjacent in the graph. Finally, we employ the diffusion distances obtained through the use of diffusion maps to regularize and constrain the predicted labels of training nodes. This regularization method is then applied to the HD-GCN training, resulting in a smoother classification surface. The model proposed in this paper effectively overcomes the limitations of information diffusion imposed only by the adjacency matrix. HD-GCN utilizes hybrid diffusion by combining information diffusion between neighborhood nodes in the feature space and adjacent nodes in the adjacency matrix. This method allows for more comprehensive information propagation among nodes, resulting in improved model performance. We evaluated the performance of HD-GCN on three well-known citation network datasets and the results showed that the proposed framework is more effective than several graph-based semi-supervised learning methods.}


\maketitle

\section{Introduction}\label{sec1}

\hspace{1.7em}
In recent years, research related to graphs has received increasing attention from researchers due to their unique structure. This is because many fields in reality have non-Euclidean data, such as social networks \cite{liben2003link,milroy2013social}, protein-protein interaction networks \cite{fout2017protein,wang2020deep,zhang2021graph}, knowledge graphs \cite{hamaguchi2017knowledge}, and other fields. Data in these domains are different from traditional Euclidean data in that it has an asymmetric and irregular structure. And describing these data by graphs can well express the structural characteristics of these data and show powerful representation. Meanwhile, graph analysis methods in machine learning can make good use of such non-Euclidean structured data and be used for tasks such as node classification, link prediction, and clustering. Graph neural networks (GNNs), a widely used graph analysis method, process graph-structured data by performing deep learning operations on the graph domain and have made remarkable progress due to their excellent performance.

While in Graph Neural Networks (GNNs), Convolutional Neural Networks (CNNs) are extended to graph-structured data by introducing Graph Convolutional Networks (GCNs). GCNs obtain convolution-like operations by aggregating information about the neighbors of a node using methods from spectral graph theory and using nonlinear activation functions to obtain low-dimensional features of graph nodes \cite{kipf2016semi}. Then, GCNs requires the entire graph to be loaded into memory for convolution during training, which is inefficient for large graph transformation work patterns. GraphSAGE \cite{hamilton2017inductive} is an inductive graph convolution method that uses sampling of some of the nodes for learning and aggregation functions to learn feature information aggregated from the neighborhoods of the nodes. It is GraphSAGE learning by sampling that makes the convolution operation does not need to load the complete graph, which is a significant improvement over GCNs in the processing of large graphs. Many different improved versions have been proposed by researchers since then \cite{chiang2019cluster,pei2020geom,yu2020forecasting}, and all have achieved promising results.

GCNs are mainly built on a semi-supervised paradigm \cite{chapelle2009semi}, where the Laplacian matrix of the graph is utilized for convolutional operations, and message aggregation is constrained by the adjacency matrix of the graph. However, it is widely recognized that both the adjacency matrix and the graph data structure are predetermined \cite{yang2016revisiting}. This approach fails to account for the manifold space, where many nodes of the graph dataset are situated closely in feature space, and this implicit neighborly relationship is not accurately captured by the adjacency matrix. As a result, the information diffusion among neighborhood nodes in feature space is not well represented, leading to suboptimal model training. Therefore, it is crucial to propose a method that enables information diffusion from the perspective of node feature space neighborhoods.

The approach of diffusion maps \cite{coifman2006diffusion} has a broad range of applications in the field of graph signal processing, where it preserves the geometry of the original graph structure and embeds the nodes as a point cloud in Euclidean space \cite{heimowitz2017unified}. It establishes similarity relations between nodes in the feature space using node information, allowing for information diffusion between neighborhood nodes through these established similarity relations. Therefore, it is reasonable to incorporate this approach into graph convolution to expand the method of diffusing node information in graph datasets.

In this paper, we introduce a Hybrid Diffusion Graph Convolutional Network (HD-GCN) framework that implements node information diffusion in two stages. Firstly, we utilize diffusion maps to facilitate the diffusion of information between nodes with similar features in the feature space. Next, the graph convolutional model is employed to facilitate the diffusion of information between adjacent nodes through the adjacency matrix. Finally, we use diffusion maps to obtain the diffusion distance of each node, and we apply regularization constraints to the HD-GCN training to make the classification surface smoother and increase the model robustness. Therefore, the proposed HD-GCN overcomes the limitations of information diffusion posed by the adjacency matrix. By combining the diffusion of information between nodes with similar features in the feature space and the diffusion of information between neighborhood nodes, the model enables a more comprehensive diffusion of node information. Traditional GCNs and its variant models can only use the adjacency matrix in the information diffusion process and do not utilize information from neighborhood nodes in the feature space. This limitation leads to poor performance on datasets with noisy data in adjacent nodes. Compared to existing GCNs, the model allows GCNs to utilize the information from adjacent nodes in the feature space, enabling more comprehensive utilization of node information. Moreover, HD-GCN eliminates some noisy features by mapping features in the diffusion maps process, enhancing GCNs noise resistance and making it more suitable for practical application scenarios.

The experiments were conducted on three publicly available citation datasets, and the results demonstrated that the proposed model outperformed most of the existing graph-based models in terms of classification performance. Compared with related work, the main contributions and advantages of this paper can be summarized as follows:

\begin{itemize}
	\item {The proposed hybrid diffusion graph convolutional model is the first to incorporate diffusion maps into GCNs, thus expanding the potential applications of diffusion maps.}
	\item {We propose a Hybrid Diffusion Graph Convolutional model that can overcome the limitations of information diffusion imposed by the adjacency matrix and can effectively leverage information from neighborhood nodes in the feature space. This model can enable more comprehensive information diffusion by utilizing information from neighborhood nodes in the feature space.}
	\item {The model utilizes diffusion maps to enable information diffusion among neighborhood nodes in the feature space, where node features are mapped to resist noise , leading to better noise resistance performance.}
\end{itemize}

The remaining sections of this paper are as follows. Section 2 provides the background knowledge for this work. Section 3 describes the proposed algorithm. The datasets, experimental setup, and results are presented in Section 4. Section 5 concludes the paper and discusses future directions for this work.

\section{Background knowledge}\label{sec2}

\hspace{1.7em}
In this section, we will examine the related work on graph-based models that are relevant to the proposed approach. The proposed model uses GCNs as the underlying classifier, so we will also provide details on how GCN is used in the approach.

The Graph Convolutional Network \cite{kipf2016semi} is based on the key idea of propagating information between nodes through information diffusion and iteratively aggregating the features of neighboring nodes using the Laplacian matrix and graph convolution. This enables the network to effectively estimate labels for unlabeled samples. The model can be represented as:
\begin{equation}
	F=f(X,A)
\end{equation}

where $F \in \mathbb{R}^{n \times d}$ denotes a label matrix representing the output values of the unlabeled data of the GCNs. $X$ denotes the feature matrix of the dataset, where $A \in \mathbb{R}^{n \times n}$ denotes the adjacency matrix associated with the samples, and the propagation law of the layers of the GCNs can be represented as:
\begin{equation}
H^{i+1}=\sigma\left(D^{-\frac{1}{2}} \tilde{A} D^{-\frac{1}{2}} H^{(i)} W^{(i)}\right)
\end{equation}

where $\tilde{A}=A+I$ denotes the $A$ matrix with added self-connections, $I$ denotes the unit matrix, and $\widetilde{D}$ denotes the degree matrix of the $\tilde{A}$ matrix. $W^{(i)}$ denotes the weight matrix corresponding to the $i$ layer of the network. $\sigma$ denotes the activation function, using the RELU function. $H^{(0)}$ is $X$. Since the GCNs model can achieve advanced results with 2-3 layers, a two-layer GCNs model is generally used. The model can be represented as:
where $A=D^{-\frac{1}{2}} \tilde{A} D^{-\frac{1}{2}}$ denotes the regularized Laplace matrix, $W^{(0)}$ denotes the input-hidden weight matrix, and $W^{(1)}$ denotes the hidden-output weight matrix. the Softmax activation function converts the output matrix by row to Each sample corresponds to the probability distribution of each category, i.e., the probability of each sample corresponding to all categories sums to 1 .

In deep neural network training, learning is performed by continuously minimizing the loss value of the estimated label and the true label. The cross-entropy function is used as the loss function in GCNs.

\begin{equation}
\mathcal{L}=-\sum_{g \in y^l} \sum_{h=1}^H Y_{g h} \ln Z_{g h}
\end{equation}

$\mathcal{L}$ denotes the cross entropy of the semi-supervised graph convolutional network; $Y^l$ denotes the set of labels corresponding to the labeled sample data in the current training data set, and $y^l$ denotes the set composed of the position ordinal numbers corresponding to each data in the training data set. In the above formula, the position number is used as the subscript of the element in $Y^l$, and $g$ is the element in the set $y^l ; Y_{g h}$ denotes the label value of the $g$ labeled sample data corresponding to category $h ; Z_{g h}$ denotes the output value of the $g$ labeled sample data corresponding to the $h$ dimension in the output of the current semi-supervised graphical convolutional network.

\section{Methods}\label{sec3}

\begin{figure}[h]
	\centering
	\includegraphics[width=0.9\textwidth]{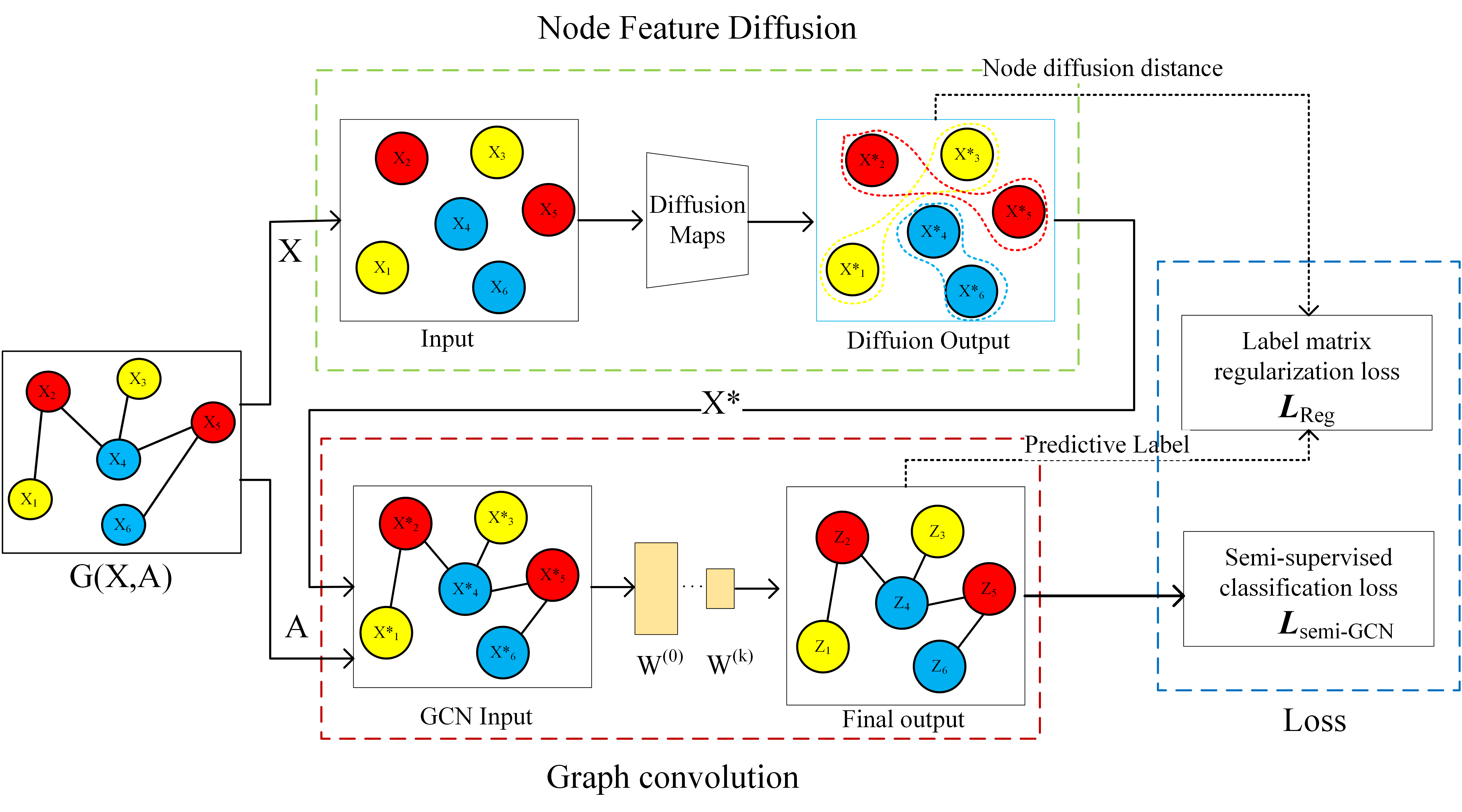}
	\caption{ Architecture of the proposed HD-GCN for Hybrid diffusion}
	\label{fig:dm-gcn}
\end{figure}

\hspace{1.7em}
In this section, we mainly describe the proposed method. We present a hybrid diffusion framework for node information by integrating diffusion maps and Graph Convolution. This innovative approach allows nodes to acquire neighborhood node information in the feature space by diffusing inter-neighborhood node information through the use of diffusion maps. Next, we apply Graph Convolution to propagate the information of adjacent nodes in the adjacency matrix, thereby enabling the nodes to capture the relationships between neighborhood nodes. Finally, the nodes are trained in a semi-supervised manner using a Graph Convolutional Network as a classifier. To regularize the predicted labels and obtain smoother final predictions, we use the diffusion distance between nodes obtained through diffusion maps to normalize the predicted labels of the classifier. The proposed hybrid diffusion process is depicted in Figure \ref{fig:dm-gcn}, wherein the node information of the dataset is initially diffused via the node feature diffusion module. Next, the diffusion node information and adjacency matrix of the dataset are further propagated through the graph convolutional module, and the HD-GCN network model is ultimately trained. We will explain each step in detail in the following sections.

\subsection{Diffusion Maps}\label{subsec2}

\hspace{1.7em}
Diffusion maps is a spectral method used for manifold learning. It achieves dimensionality reduction by re-organizing data based on the geometric parameters of the underlying data structure \cite{hu2020open}

Denote the dataset node information as $X, X \in R^{n \times d}, n$ denotes the number of samples in the dataset and $d$ denotes the feature dimension of the samples. Diffusion maps exploits the connectivity of a dataset using local similarity as a metric for creating a time-based diffusion process. The connectivity between two data points can be defined as the probability of random walk from one point to another, which is expressed by the kernel function of these two data points: $k: X \times X \rightarrow R$. The diffusion kernel function satisfies the following properties:
\begin{itemize}
	\item {$k$ is symmetric: $k(x, y)=k(y, x)$}
	\item {$k$ is positivity preserving: $k(x, y) \geq 0$}
\end{itemize}

The kernel function in diffusion maps serves as a representation of the connectivity between data points in the dataset $X$, and thus defines the local geometric structure of the data. Choosing the right kernel function is crucial, as it captures the unique characteristics of the dataset. Typically, the Gaussian kernel function is selected to measure the similarity between data points:

\begin{equation}
k(x, y)=\exp \left(-\frac{\|x-y\|^2}{2 \sigma^2}\right)
\end{equation}

The connectivity between data points is determined by the Gaussian kernel function, which is selected based on the specific characteristics of the dataset. The variance of the Gaussian kernel function is denoted by $\sigma$, and a larger $\sigma$ value indicates weaker connectivity between data points.
With the given kernel function $k(x, y)$, we can obtain the similarity matrix $S$ for the construction of the data set $\mathrm{X}$
\begin{equation}
S_{i j}=k\left(x_i, x_j\right) \quad x_i, x_j \in X
\end{equation}

Then the similarity matrix $S$ is normalized so that the sum of each row of it is 1 . This gives us a normalized matrix $P=P^{(1)}=\left\{p_{i j}^{(1)}\right\}$, where the elements are: $p_{i j}^{(1)}=\frac{s_{i j}}{\sum_z s_{i z}}$. This normalized matrix $P$ can be interpreted as a state transfer matrix $M$ over the data set $X$. This matrix represents the probability of random walk between individual data points. In other words, the state transfer matrix $M$ in diffusion maps represents the probability of transitioning from one data point to another after one step of random walk. The element $M_{i j}$ in the matrix represents the probability of transitioning from point $i$ to point $j$, and these probabilities are typically calculated based on the similarity between the data points. And if after $t$ steps of random walk, the corresponding transfer probability $M^t=(M)^t$. When we calculate the state transfer matrix $M^t$ for increasing values of $t$, we can observe the dataset at different scales. This is the feature diffusion process in the manifold space, and we see that the local connectivity is integrated to provide the overall connectivity of the dataset. In this diffusion process, the nodes of the dataset and its neighborhood nodes make information aggregation.

Calculating the $k$ maximum eigenvalues and corresponding eigenvectors of the corresponding state transfer matrix $M^t$ after $t$ steps of random walk, we thus obtain: $M^t \psi(x)=$ $\lambda \psi(x)$. From this, we know that the corresponding feature vector can represent a new set of coordinates of the dataset in the feature space, and we obtain the diffusion maps from the original data to the k-dimensional space embedded in the original space as:
\begin{equation}
\Psi^{(t)}(x)=\sum_k \lambda_k^t \psi_k(x)
\end{equation}

$\Psi^{(t)}(x)$ denotes the node information after the diffusion process through diffusion maps, where each node's features already contain the information of the neighborhood nodes in its feature space.

\subsection{Hybrid diffusion method}\label{subsec3}

\hspace{1.7em}
The node feature of the dataset is denoted as $X=\left[x_1, x_2, \ldots, x_n\right], X \in R^{n \times d}, n$ denotes the number of samples in the dataset and $d$ denotes the number of feature dimensions of the samples. To achieve information diffusion between feature space neighborhood nodes of the dataset, we perform diffusion maps on the dataset, which enables the node feature information $X$ to diffuse between neighborhood nodes in the feature space. The node information after the diffusion of information among the nodes in the feature space neighborhood is:

\begin{equation}
\Psi^{(t)}\left(x_i\right)=\sum_k \lambda_k^t \psi_k\left(x_i\right)
\end{equation}

where $t$ is the number of diffusion times, $\lambda_k$ denotes the $k$ eigenvalue of the $i$ node feature after eigendecomposition, and $\psi_k\left(x_i\right)$ is the $k$ eigenvector of the $i$ node feature after eigendecomposition. $\Psi^{(t)}\left(x_i\right)$ is the $i$ node feature information of the dataset after $\mathrm{t}$ diffusion times of diffusion maps.

The dataset is then diffused by the Graph Convolution model to diffuse the information of the adjacency relations. The resulting GCNs model based on node feature information diffusion:
\begin{equation}
F_{H D-G C N}=\operatorname{softmax}\left(\breve{A} \sigma\left(\breve{A} W^{(0)} \Psi^{(t)}\left(x_i\right)\right) W^{(1)}\right)
\end{equation}

where $\breve{A}$ denotes the regularized Laplace matrix, $W^{(0)}$ denotes the input-hidden weight matrix, and $W^{(1)}$ denotes the hidden-output weight matrix. The SoftMax activation function converts the output matrix by row so that each sample corresponds to a probability distribution across all categories, where the probabilities of each category for a given sample sum up to 1 .

$F_{H D-G C N}$ diffuses the node information after diffusion maps by Graph convolution operation to diffuse the node information among adjacency nodes, to achieve the purpose of node information hybrid diffusion.

In diffusion maps, the diffusion distance between two points at time $t$ can be measured as the similarity of two points in the observation space with the connectivity between them. It is given by

\begin{equation}
	D_{i j}=\sum_k \lambda_k^{2 t}\left(\psi_k\left(x_i\right)-\psi_k\left(x_j\right)\right)^2
\end{equation}

The diffusion distance maintains the original distribution of the dataset in the feature space. This distance is robust to noise because the distance between two points depends on all possible paths of length $\mathrm{t}$ between the two points. While GCNs only considers the neighboring relationships between nodes in the adjacency matrix, the diffusion maps approach takes into account the similarity of node features in the feature space. This means that even nodes without adjacency relationships in the adjacency matrix can still be connected through neighbor relationships in the feature space using the approach. By preserving neighbor information in the feature space, the method can better capture the local geometric structure of the dataset. However, the feature representations of each node change after information diffusion through the adjacency matrix, and thus their relationships in the feature space also change correspondingly. We aim to preserve the relationships between nodes in the feature space using diffusion distance. To achieve this goal, we use a combination of cross-entropy loss and node feature similarity regularization loss to learn a GCNs model based on hybrid diffusion. This approach helps to maintain the original distribution of the dataset while preserving its neighbor relationships in the feature space. The loss function used is:

\begin{equation}
\mathcal{L}=-\sum_{i=1}^L \sum_{k=1}^c y_{i k} \log \left(F_{i k}\right)+\alpha \sum_{i=1}^N \sum_{j=1}^N\left\|F_{i *}-F_{j *}\right\|^2 D_{i j} \cdot A_{i j}
\end{equation}

where $F_{i *}$ denotes the $i$ row in the label matrix $F $. $y_{i k}$ denotes the true label distribution of the $i$ sample. $L$ is the number of labeled samples. $\alpha$ is the balance parameter that balances these two terms. $A=[A_{i j}]$ is the adjacency matrix of the dataset, where $ A_{i j}$ represents the adjacency relationship between sample $i$ and sample $j$; $D=[D_{i j}]$ is the diffusion distance obtained by diffusion maps, where $D_{i j}$ represents the diffusion distance between the $i$ sample and the $j$ sample; The first term is a cross-entropy loss, while the second term is a regular term for the similarity of the node prediction labels. Thus, minimizing the first term ensures that the estimated labels of the labeled nodes are as close as possible to their true labels. On the other hand, by minimizing the second term. we will reduce the node prediction label noise and make the classification surface of the model smoother. Obviously, if a pair of nodes are highly similar in the feature space and also adjacent in the adjacency matrix, they will generate two highly similar predicted label vectors $F_{i *}$ and $F_{j *}$, which would also minimize the second term $\left\|F_{i *}-F_{j *}\right\|^2 D_{i j} \cdot A_{i j}$ in the equation(10).

Hence, we incorporate the similarity matrix of sample node features obtained from diffusion maps into the loss function as a regularization term to reduce sample noise, ensure a smoother classification surface, and minimize the risk of overfitting.

\section{Experiments}\label{sec4}

\hspace{1.7em}
In this section, we presented the experimental results of HD-GCN on three datasets and compared it with several graph-based semi-supervised classification methods. We also demonstrated that the HD-GCN model has better robustness than some graph-based semi-supervised classification methods in graph node classification tasks.

\subsection{Datasets}\label{subsec4}

\hspace{1.7em}
The predictive power of the model was evaluated on three citation network datasets, Cora, Citeseer and Pubmed \cite{sen2008collective}, which have been used as benchmarks for many graph-based semi-supervised classification tasks. The dataset statistics are shown in Table \ref{tab1} and the dataset is briefly described as follows:

\begin{itemize}
	\item {\textbf{Cora}: The dataset consists of 2708 scientific publications, each described by a 1433-dimensional word vector with values of 0 and 1, representing whether the corresponding word appears in the paper or not. Cora's publications are classified into 7 categories.}
	\item {\textbf{Citeseer}: This dataset uses a similar representation to Cora and consists of 3327 documents, publications are classified into 6 categories and the sample is described by a 3703-dimensional word vector.}
	\item {\textbf{Pubmed}: This dataset consists of 19717 scientific publications on diabetes in the Pubmed database, with publications classified into three categories and described by a TF/IDF-weighted word vector in a dictionary of 500 unique words.}
\end{itemize}

\begin{table}[h]
	\tabcolsep=0.75cm
	\caption{Citation network datasets statistics [13].}\label{tab1}%
	\begin{tabular}{@{}lllll@{}}
		\toprule
		Dataset & Nodes & Edges & Classes & Features\\
		\midrule
		Cora    & 2708   & 5429  & 7 & 1433  \\
		Citeseer    & 3327   & 4732  & 6 & 3703  \\
		Pubmed    & 19717   & 44338  & 3 & 500  \\
		\botrule
	\end{tabular}
\end{table}

The performance of graph-based algorithms heavily depends on the adjacency matrix. In the case of the citation network dataset utilized in this paper, the adjacency matrix is constructed based on whether two nodes (i.e., papers) cite each other.

\subsection{Experimental setup}\label{subsec5}

\hspace{1.7em}
We compared the proposed HD-GCN model with some traditional machine-learning methods and some advanced graph-based methods. These methods can be divided into two categories: (1) traditional machine learning algorithms, and (2) graph-based convolutional networks.

The traditional machine learning algorithms we considered in this study include the multilayer perceptron (MLP)\cite{rumelhart1986learning} and support vector machine (SVM)\cite{cortes1995support}. The graph-based models compared in this study include the representative semi-supervised graph convolution networks (GCNs) \cite{kipf2016semi}, Graph Attention Network (GAT) \cite{velivckovic2017graph}, Topology Adaptive Convolutional Network (TAGCN) \cite{du2017topology}, Predict then Propagate: Graph Neural Networks meet Personalized PageRank (APPNP) \cite{gasteiger2018predict}, and Attention-based Graph Neural Network for Semi-supervised Learning (AGNN) \cite{thekumparampil2018attention}.

All experiments in this paper were implemented on the Pytorch framework. The graph-based approach is implemented through Pytorch Geometric (PYG), a geometry learning extension library based on the Pytorch framework. The traditional machine learning methods were implemented using the Scikit-learn package. 

Adam was used as an optimizer to train all the above graph-based models. During the training phase, the hyperparameters and network configuration of each model follow the default settings provided in PYG as a benchmark. The learning rate for all models was set to 0.01 and the dropout rate was defined as 0.5, except for GAT which used a dropout rate of 0.6. To ensure fairness in comparison, MLP and SVM were trained using the default settings in Scikit-learn, with the maximum iteration limit for MLP set to 1000 to guarantee complete convergence.

The proposed method has one extra hyperparameter: . We set this parameter to a subset of values belonging to $\left\{1,10^{-1}, 10^{-2}, 10^{-3}, 10^{-4}, 10^{-5}\right\}$.

\subsection{Method comparison}\label{subsec6}

\hspace{1.7em}
In order to evaluate the model more comprehensively, we use multiple benchmark models to compare the effects, and the experimental results are shown in Table \ref{tab2}.
\begin{table}[h]
	\tabcolsep=1cm
	\caption{Results of multiple citation network dataset.}\label{tab2}%
	\begin{tabular}{@{}llll@{}}
		\toprule
		Method & Cora & Citeseer & Pubmed \\
		\midrule
		SVM \cite{cortes1995support} & 0.555 & 0.584 & 0.714 \\
		MLP \cite{rumelhart1986learning} & 0.527 & 0.498 & 0.698 \\
		GCN \cite{kipf2016semi} & 0.818 & 0.701 & 0.799 \\
		SGC \cite{wu2019simplifying} & 0.786 & 0.726 & 0.762 \\
		GAT \cite{velivckovic2017graph} & 0.817 & 0.731 & 0.775 \\
		AGNN \cite{thekumparampil2018attention} & 0.822 & 0.699 & 0.792 \\
		TAGCN \cite{du2017topology} & 0.830 & 0.707 & 0.792 \\
		APPNP \cite{gasteiger2018predict} & 0.829 & 0.702 & 0.804 \\
		\midrule
		\textbf{HD-GCN} & $\mathbf{0 . 8 3 6}$ & $\mathbf{0 . 7 4 2}$ & $\mathbf{0 . 8 0 0}$ \\
		\textbf{Reg-HD-GCN} & $\mathbf{0 . 8 3 8}$ & $\mathbf{0 . 7 4 5}$ & $\mathbf{0 . 8 0 0}$ \\
		\botrule
	\end{tabular}
\end{table}

From Table \ref{tab2}, we can see that the accuracy of the hybrid diffusion model has improved in comparison with the majority of models. On the Cora dataset, the method shows a significant improvement compared to all other methods. On the Citeseer dataset, the method has the most significant improvement, with an effective improvement compared to all other methods. In the Pubmed dataset, the method is only less effective than that of APPNP. It can also be observed that the regular term has an effect enhancing effect on both Cora and Citeseer datasets. Also, the inclusion of the regularization constraint has an improvement on the prediction results except for the Pubmed dataset. This may be because the Pubmed dataset is sparser compared to the other two datasets, and most of the nodes do not have any adjacency relationships, which leads to poor performance of the regularization constraint.

\subsection{Parameter analysis}\label{subsec8}

\hspace{1.7em}
In the hybrid diffusion method, diffusion maps diffuse information between nodes in feature space neighborhoods. The main parameters that affect the diffusion maps method are diffusion time.

\begin{figure}
	\centering
	\includegraphics[width=0.7\linewidth]{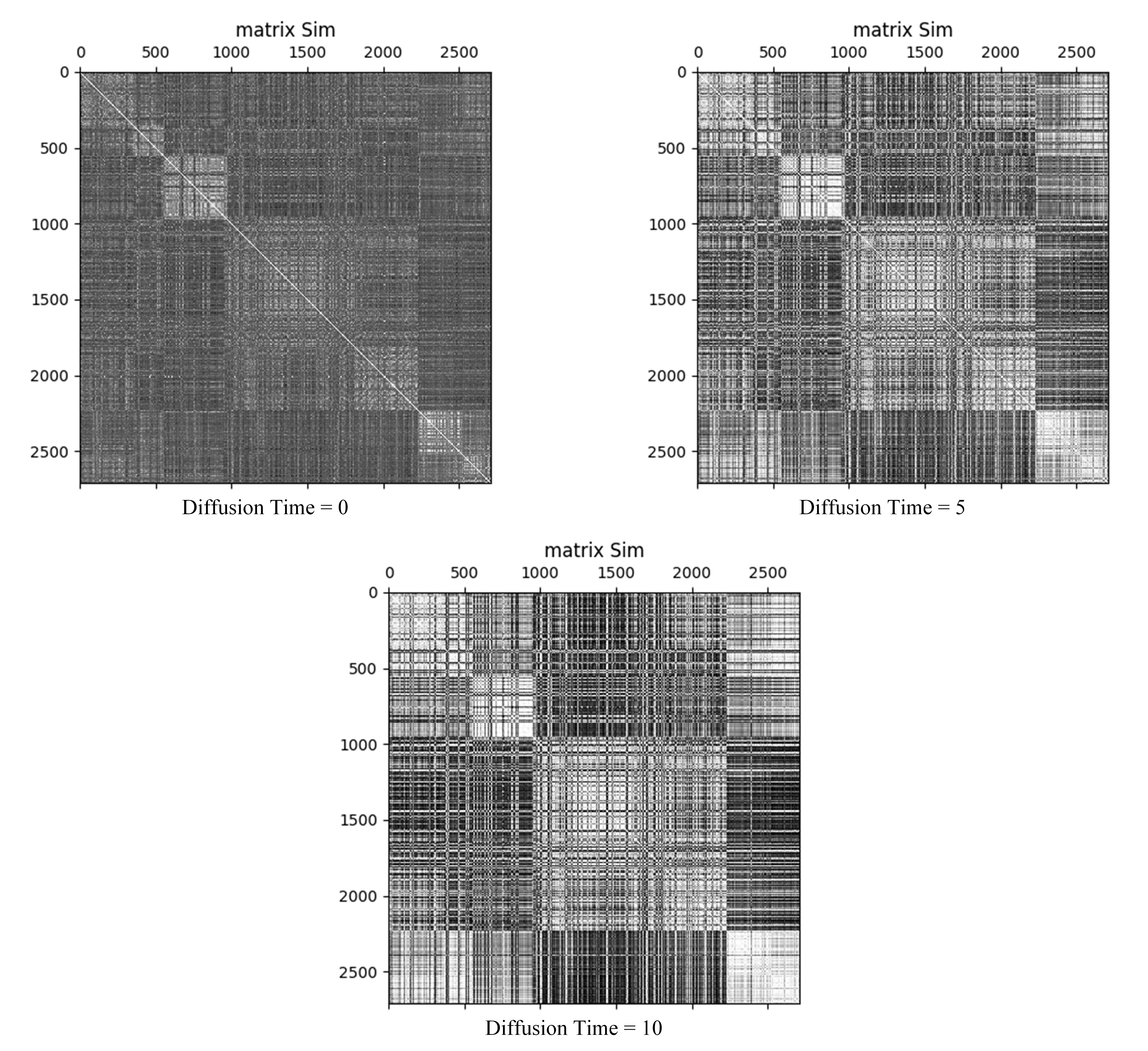}
	\caption{ The effect of diffusion time on the similarity of node features. The magnitude of similarity is expressed by the brightness of the pixel points, and the brighter the pixel points are, the greater the similarity of the two nodes. From left to right, the similarity of nodes with diffusion counts of 0, 5, and 10 are shown respectively.}
	\label{fig:diffusiontime}
\end{figure}

To demonstrate the impact of the node feature space diffusion module on the diffusion of node information, we used the Cora dataset and processed it with diffusion maps.  The 2,708 nodes with the same category in the dataset were reclassified together. The degree of similarity between these nodes was demonstrated using the cosine similarity function, which was used to explain the effectiveness of information diffusion based on the similarity between the nodes. Figure \ref{fig:diffusiontime} shows the changes in node feature similarity under different diffusion times. From Figure \ref{fig:diffusiontime}, it can be observed that as the number of diffusion steps increases, the similarity between nodes of the same category gradually increases.

\begin{figure}
	\centering
	\includegraphics[width=0.7\linewidth]{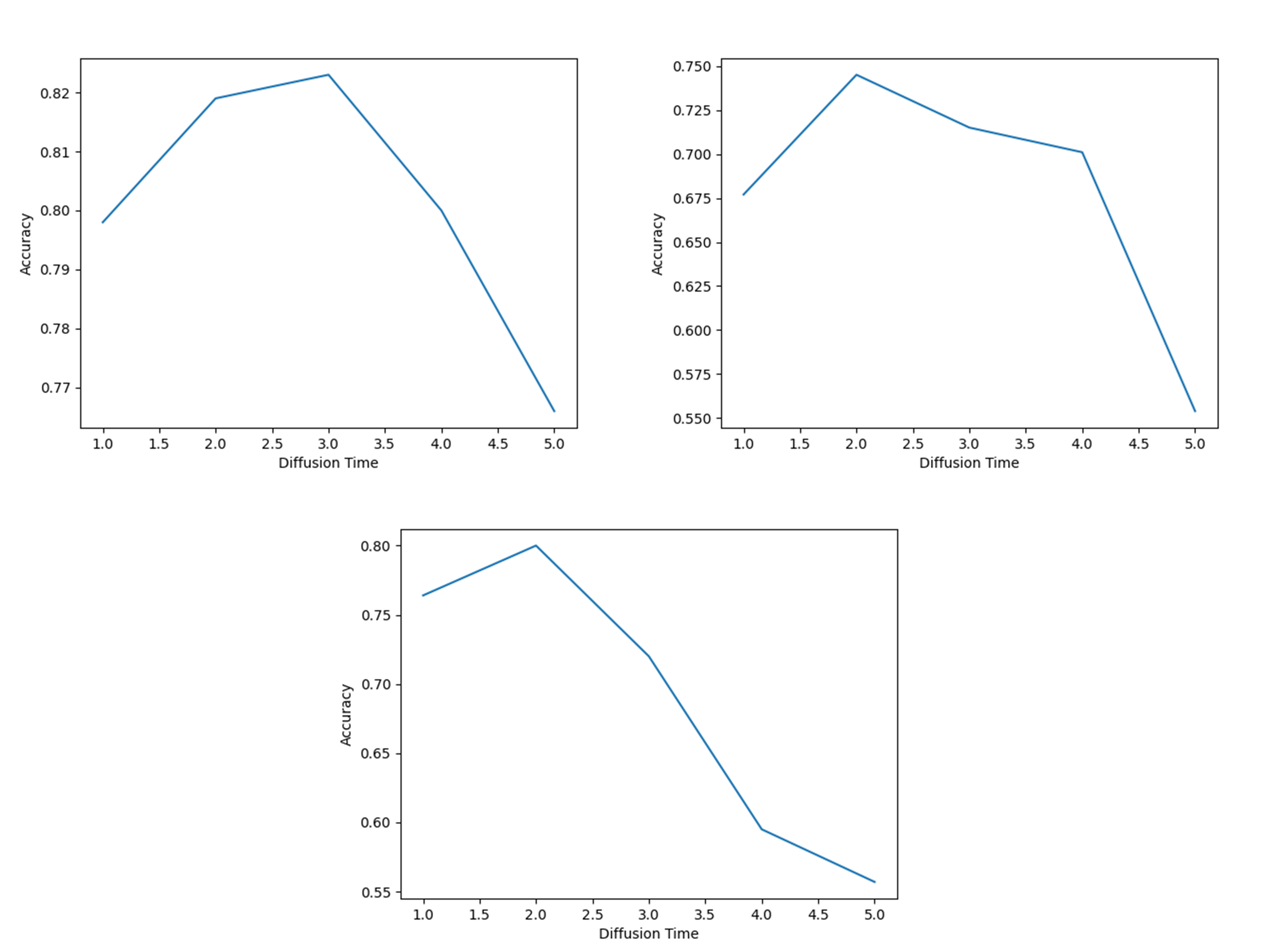}
	\caption{The effect of diffusion time on the accuracy of each dataset in the hybrid diffusion method, from left to right, (a) Cora, (b) Citeseer, (c) Pubmed, respectively.}
	\label{fig:dt}
\end{figure}

Figure \ref{fig:dt} demonstrates the impact of the diffusion time parameter in diffusion maps on node classification accuracy. From Figure \ref{fig:diffusiontime} and \ref{fig:dt}, we can observe that even though the similarity between nodes of the same category increases with the rise in diffusion time, the training performance does not necessarily improve with more diffusion steps. In fact, the trend shows an increase followed by a decrease in accuracy. The reason for this is that as the diffusion time increases, the similarity between node features also increases. When the diffusion time is small, the feature information of nodes and their similar nodes in the feature space is not fully diffused, which leads to a low classification accuracy for some nodes. On the other hand, when the diffusion time is large, the feature information of nodes and their similar nodes in the feature space is excessively diffused, resulting in the loss of differences between nodes during network training and decreasing training accuracy.

\subsection{The effect of noise}\label{subsec9}

\hspace{1.7em}
During the generation of graph datasets, some inaccurate data may be produced, yet many graphs neural networks fail to consider the presence of such cases. As an example, practitioners often resort to inexpensive alternatives, like manual and automated label generation through cooperation \cite{hu2020open}, which inevitably leads to samples with erroneous labels. As neural networks, including GNNs, are capable of memorizing any (random) label \cite{pan2016tri,zhang2021understanding,arpit2017closer}, such noisy labels can impede their ability to generalize effectively. The node feature composition of the citation dataset we used (Cora, Citeseer, Pubmed) \cite{sen2008collective} is determined by the use of specific words, which is highly likely to generate imprecise features during the generation process, and these inaccuracies can also impact the final performance of the network's generalization ability. 

Each node in the citation dataset represents a research paper, which is represented by a specific dimension word vector. Thus, each node possesses a specific number of features equal to the dimension of the word vector. Each element of the word vector corresponds to a word and takes on only two possible values, 0 or 1. A value of 0 indicates that the corresponding word is not present in the paper, while a value of 1 indicates that the word is present in the paper. Noise can be added to the dataset by introducing word features that do not exist in the original paper in the word vector, or by causing word features that exist in the original paper to be lost.

\begin{figure}
	\centering	\includegraphics[width=0.7\linewidth]{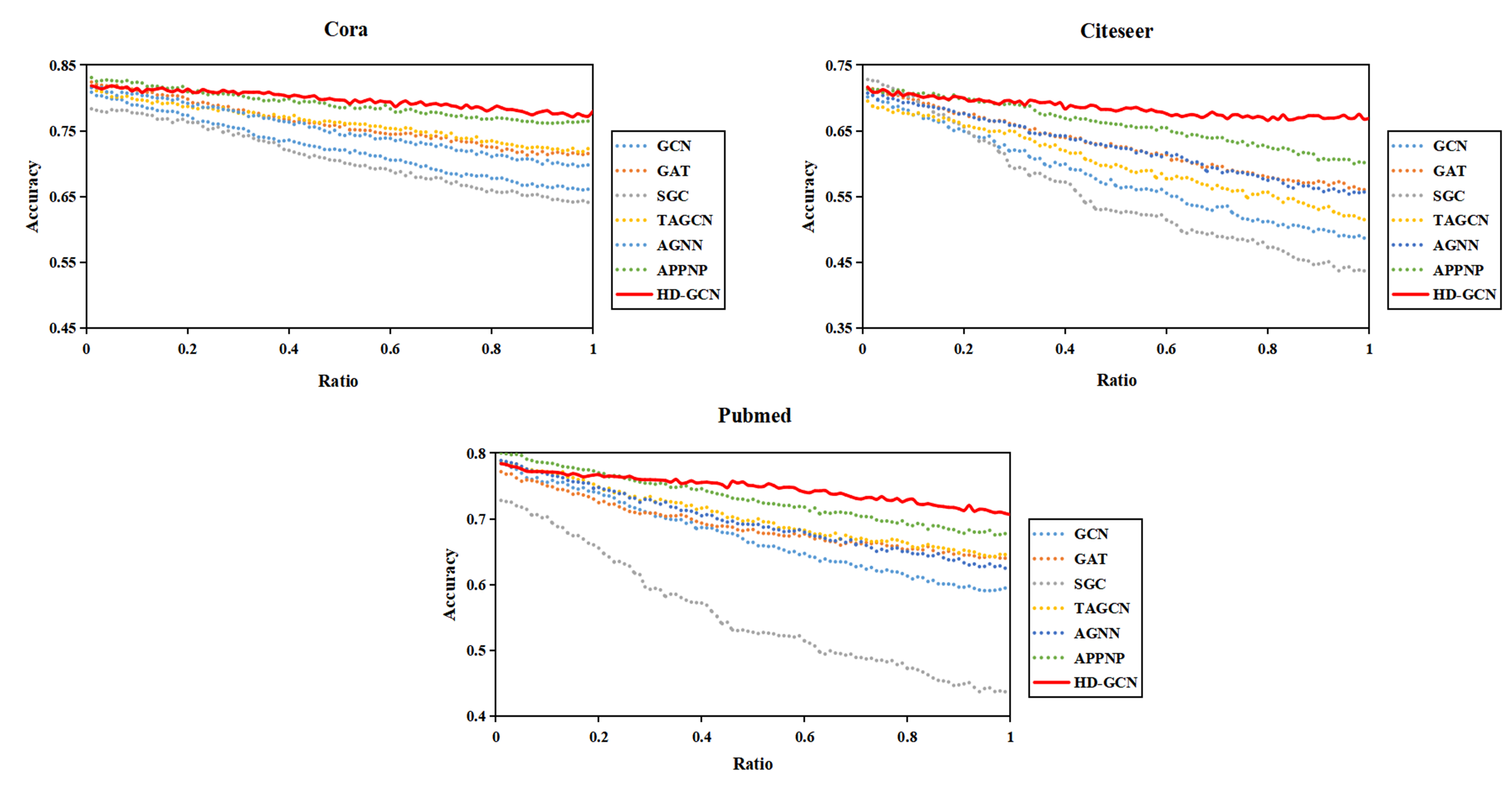}
	\caption{Classification accuracy (\%) of the proposed method for different ratios of noise (a) Cora dataset. (b) Citeseer dataset. (c) Pubmed dataset.}
	\label{fig:noise}
\end{figure}

Since noise of feature could impact the accuracy of model, it is interesting to analyze the robustness of different models to noise. One way to achieve this is by adding or removing noise in the node features of the dataset. In this section, we will adjust the proportion of noise in the dataset to investigate the robustness of different models to noise. In Figure \ref{fig:noise}, the horizontal axis represents the ratio of noise to the significant bit of features in the dataset, and the vertical axis represents the classification accuracy of the model.

The proposed method demonstrates superior noise robustness compared to other graph-based semi-supervised classification methods, as shown in Figure \ref{fig:noise}, with better classification accuracy under most noise ratios. This can be attributed to the use of diffusion maps to facilitate information diffusion between nodes in the feature space, allowing each node to diffuse its information to similar nodes in the manifold space and thus resist noise.

\section{CONCLUSION}\label{sec5}

\hspace{1.7em}
We have proposed a hybrid diffusion model based on diffusion maps. This model employs diffusion maps to diffuse information on neighbor relationships in the feature space of nodes, followed by graph convolution to diffuse information on adjacent nodes, to handle datasets that have undergone diffusion of information in the node feature space. The diffusion maps method enables the diffusion of information between neighborhood nodes in the feature space of the nodes. The method of node information diffusion is extended to hybrid diffusion by combining the information diffusion between neighborhood nodes in the feature space and the information diffusion between adjacent nodes in the adjacency matrix. These two methods of diffusion of information complement each other and promote comprehensive learning of similar node information in the model. Furthermore, during training, a regular term based on feature similarity of node is used to preserve the relationship between nodes in the feature space, resulting in a smoother classification surface for the model. This model employs a neural network structure similar to that of traditional graph convolutional networks. Therefore, training and testing complexity are almost equivalent, but by implementing a hybrid diffusion framework, it enables information propagation between neighborhood nodes in feature space, surpassing the limitation of the adjacency matrix on information diffusion and expanding the ways in which information is propagated, compared to traditional GCNs.

In the future work, we will focus on the following directions: (1) Utilizing the node features at different diffusion times more comprehensively, as the node features at different diffusion times can reveal features at different scales, thereby allowing the final graph embedding to be learned more effectively. (2) Reducing the model's time complexity is crucial in practical application scene.

\bibliography{hyperref}

\end{document}